\documentclass[11pt]{article}

\usepackage[final]{acl}

\usepackage{times}
\usepackage{latexsym}
\usepackage{amsmath}
\usepackage{multirow}
\usepackage{booktabs}
\usepackage[T1]{fontenc}

\usepackage[utf8]{inputenc}

\usepackage{microtype}

\usepackage{inconsolata}

\usepackage{graphicx}

\title{Beyond Policy Optimization: A Data Curation Flywheel for \\Sparse-Reward Long-Horizon Planning}

\author{\normalsize
  \textbf{Yutong Wang}\textsuperscript{1}\thanks{\ \ Equal contribution.},
  \textbf{Pengliang Ji}\textsuperscript{2}\footnotemark[1],
  \textbf{Kaixin Li}\textsuperscript{3},
  \textbf{Baolong Bi}\textsuperscript{4},
  \textbf{Tao Feng}\textsuperscript{5},
  \textbf{Guillaume Sartoretti}\textsuperscript{1}\thanks{\ \ Corresponding author.}
\\[2mm]
  \parbox{0.95\linewidth}{\centering
    \textsuperscript{1}Department of Mechanical Engineering, National University of Singapore \\
    \textsuperscript{2}Robotics Institute, Carnegie Mellon University \\
    \textsuperscript{3}School of Computing, National University of Singapore \\
    \textsuperscript{4}Institute of Computing Technology, Chinese Academy of Sciences \\
    \textsuperscript{5}Department of Computer Science and Technology, Tsinghua University
  }
\\
  \small{
    \href{mailto:e0576114@u.nus.edu}{e0576114@u.nus.edu},
    \href{mailto:guillaume.sartoretti@nus.edu.sg}{guillaume.sartoretti@nus.edu.sg}
  }
}

\begin{document}
\maketitle
\begin{abstract}
Large Language Reasoning Models have demonstrated remarkable success on static tasks, yet their application to multi-round agentic planning in interactive environments faces two fundamental challenges. First, the intractable credit assignment problem renders conventional reinforcement learning ineffective in sparse-reward settings. Second, the computational overhead of verbose, step-by-step reasoning histories is prohibitive. To address these challenges, we propose BPO, a three-stage framework (bootstrapping, extrapolation, and refinement) that establishes a self-improving data flywheel to develop robust reasoning models for long-horizon, sparse-reward environments. Our framework first bootstraps efficient reasoning using the proposed planning quaternions with long-short chain-of-thought fusion. It then extrapolates to out-of-distribution tasks through complexity-stratified curriculum learning. Finally, the model iteratively refines itself by learning exclusively on experiences selected via reward-gated rejection sampling. Experiments on ALFWorld, ScienceWorld, and WebShop demonstrate that our approach achieves state-of-the-art with significant token efficiency, providing a new recipe for reasoning models in agentic planning.
\end{abstract}

\section{Introduction}

Recent advances in Large Language Models (LLMs), particularly the emergence of models capable of ``System 2'' reasoning~\cite{jaech2024openai, deepseekr1}, have marked a paradigm shift~\cite{ke2025survey}. By generating explicit chains of thought, these models achieve expert-level performance in static, self-contained domains such as mathematics and programming~\cite{wei2025swe, chen2025evaluating}. They are inherently better suited for agentic planning, bringing a global perceptual scope that enables multi-step foresight and planning with subgoals. This raises a critical question: can this powerful reasoning paradigm be extended to dynamic, interactive environments to build autonomous agents capable of long-horizon planning?~\cite{shinn2024reflexion}. However, integrating explicit reasoning into agentic tasks presents two major challenges.

The first challenge is the intractable credit assignment problem endemic to long-horizon tasks with sparse rewards~\cite{shridhar2020alfworld, wang2022scienceworld}. In many interactive settings, rewards, typically binary success indicators, are delivered only after a long and complex sequence of actions. This significant temporal delay renders the feedback signal too weak for reinforcement learning algorithms to effectively propagate credit back through the trajectory. As our experiments show, even advanced preference optimization algorithms, such as Group Relative Policy Optimization (GRPO)~\cite{shao2024deepseekmath}, yield only marginal improvements, highlighting the fundamental limitations of optimizing reasoning LLMs in this regime.
The second challenge is the computational and cognitive overhead of verbose reasoning. While detailed reasoning is crucial for complex problem-solving, its naive inclusion at every step of an interactive trajectory inflates context length, incurs substantial computational costs, and risks displacing critical information from earlier in the trajectory out of the model's limited attention window~\cite{bai2023longbench}.

To address the challenges of adapting reasoning LLMs to long-horizon, sparse-reward planning settings, this work introduces BPO, a novel three-stage framework of bootstrapping, extrapolation, and refinement that establishes a self-improving \textit{data flywheel}. First, our framework bootstraps efficient reasoning through Planning Quaternion Synthesis, a specialized data generation process that disentangles verbose deliberation from concise planning intent. This stage is paired with a long-short chain-of-thread fusion strategy that not only ensures contextual efficiency but also mirrors a plausible cognitive model of long-term planning, where detailed immediate reasoning is compressed into high-level conclusions for future reference. Second, the agent's robustness is cultivated via Curriculum Synthesis, which proactively generates challenging out-of-distribution tasks and organizes them into a complexity-stratified curriculum to systematically expand its capabilities. Finally, the agent achieves self-refinement through reward-based trajectory improvement. In this stage, the sparse environmental reward acts as a deterministic criterion for success-contingent rejection sampling, creating a virtuous cycle of iterative fine-tuning on exclusively successful trajectories.

In summary, our contributions are threefold:
\begin{itemize}
    \item We introduce BPO, a three-stage framework of bootstrapping, extrapolation, and refinement for reasoning LLMs in sparse-reward, long-horizon agentic tasks.
    \item We propose a suite of techniques, including planning quaternion data structure, long-short chain-of-thought fusion, and curriculum synthesis, collectively enable robust, efficient, and generalizable reasoning.
    \item Extensive experiments on ALFWorld, ScienceWorld, and WebShop show our approach achieves state-of-the-art performance, surpassing fine-tuning and larger proprietary models with significant fewer tokens.
\end{itemize}

\section{Task Formulation}

We formalize the problem of a reasoning LLM agent interacting with an environment as a \textit{partially observable Markov decision process (POMDP)}, augmented to explicitly account for the agent's internal reasoning process. The environment is defined by the tuple
\[
\mathcal{M} = (U, S, A, O, E, T, R),
\]
where $U$ represents the space of natural language task instructions, $S$ denotes the latent space of environment states, $A$ is the discrete set of permissible actions, $O$ describes the space of observations provided by the environment, and $E$ refers to the space of explicit reasoning traces (i.e., thoughts). The deterministic state transition function is
\[
T : S \times A \rightarrow S,
\]
and the sparse reward function is
\[
R : S \times A \rightarrow \{0,1\}.
\]

At each time step \(t\), an agent parameterized by \(\theta\) receives an observation \(o_t\). Conditioned on the instruction \(u\) and the interaction history, its policy \(\pi_\theta\) generates a reasoning trace \(e_t\) and then selects an action \(a_t\). The resulting generative process is:

\[
\begin{aligned}
[e_t, a_t] \sim \pi_\theta(&u, (o_1, e_1, a_1), \dots, \\
                           &(o_{t-1}, e_{t-1}, a_{t-1}), o_t).
\end{aligned}
\]

The execution of $a_t$ transitions the environment to a new state $s_{t+1} = T(s_t, a_t)$, yielding a new observation $o_{t+1}$. This interaction repeats until a terminal state or a maximum number of steps is reached. The agent's objective is to learn a policy $\pi_\theta$ that maximizes the expected cumulative reward.

For supervised fine-tuning, the learning objective is to maximize the likelihood of generating the correct reasoning and action from expert trajectory. A trajectory \(\mathcal{T}\) consists of a sequence of historical states leading to a target reasoning–action pair. Given the instruction and the history of observation–thought–action tuples, the model is trained to predict the subsequent thought $e_N$ and action $a_N$:
\begin{equation} \label{eq:trajectory}
\begin{aligned}
\mathcal{T} = [u,\ &(o_1, e_1, a_1), \dots, (o_{N-1}, e_{N-1}, a_{N-1}),\\
              &o_N \rightarrow (e_N, a_N)].
\end{aligned}
\end{equation}

This formulation unifies the instruction, historical context, and reasoning into a single sequence, enabling the training of reasoning LLM agents.

\begin{figure*}[t]
\centering
\includegraphics[width=\textwidth]{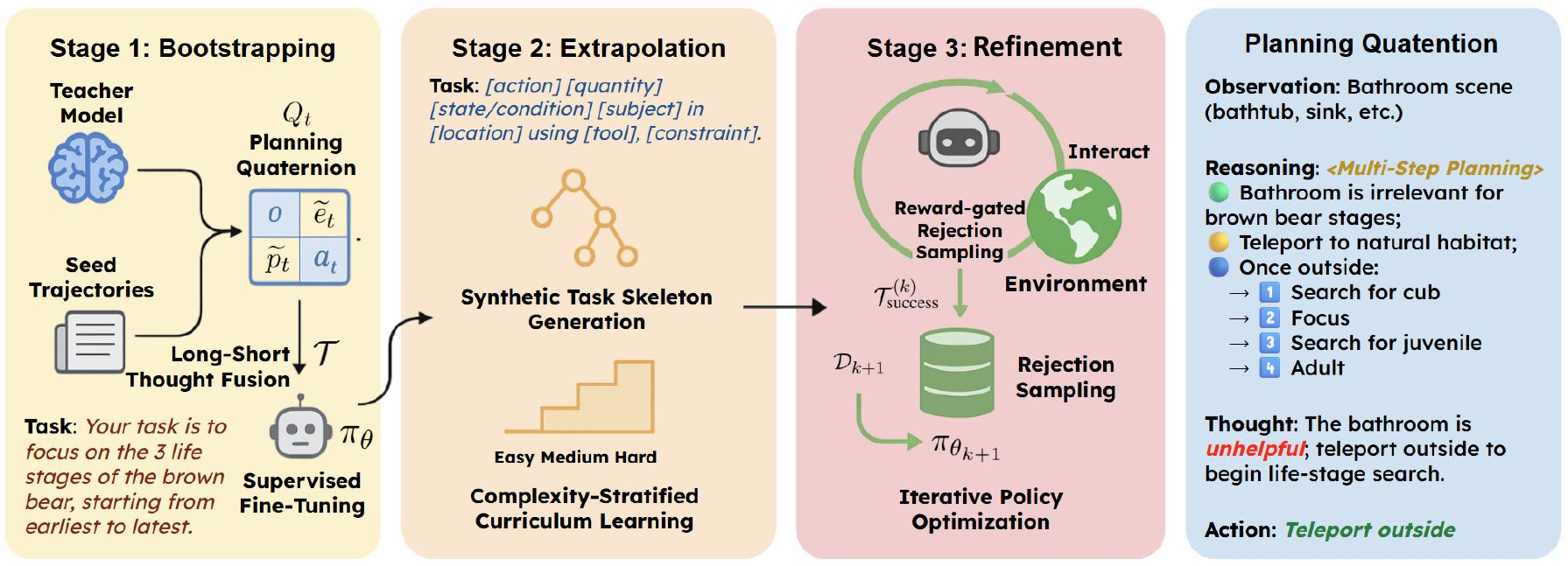}
\caption{Overview of our three-stage framework for training reasoning LLMs in long-horizon, sparse-reward environments. \textbf{Stage 1: Bootstrapping Reasoning via Planning Quaternion}. An initial agent is trained via supervised fine-tuning on planning quaternion data from seed trajectories using a teacher model. \textbf{Stage 2: Extrapolation via Curriculum Synthesis}. The agent's capabilities are expanded by training on a curriculum of synthetically generated, out-of-distribution tasks of increasing complexity. \textbf{Stage 3: Self-Refinement via Reward-Gated Refinement}. The agent enters an iterative self-improvement loop, generating new trajectories and enriching the dataset for subsequent fine-tuning through rejection sampling while interacting with the environment.}
\label{fig:pipeline}
\end{figure*}

\section{Method}
Our proposed framework is designed to train LLM agents capable of effective long-horizon planning under sparse rewards. The methodology deliberately circumvents the intractable credit assignment problem that plagues conventional reinforcement learning by re-purposing the sparse environmental rewards, from a noisy learning signal, into an unambiguous filter for data curation. As shown in Figure~\ref{fig:pipeline}, we propose a three-stage pipeline: (1) bootstrapping a foundation of efficient reasoning, (2) extrapolating via a synthetic curriculum, and (3) achieving perfection through iterative refinement guided by sparse rewards.

\subsection{Stage 1: Bootstrapping Reasoning via Planning Quaternion Synthesis}
The initial stage addresses the dual challenges of acquiring high-quality, multi-faceted reasoning data and managing the computational overhead associated with verbose thought processes.

\subsubsection{The Planning Quaternion Data Flywheel}
High-quality data that jointly models detailed deliberation and concise, actionable planning is scarce. To overcome this limitation, we introduce the \textit{Planning Quaternion Data Flywheel} to systematically generate rich training data. The process unfolds in four steps:

\textbf{Seed Trajectory Curation:} We begin by collecting a set of successful trajectories from a powerful, off-the-shelf reasoning model (e.g., DeepSeek-R1) in the target environment. These trajectories provide initial examples of observation-reasoning-action sequences.

\textbf{Synthetic Expansion:} Using these seed trajectories as few-shot exemplars, we prompt a highly capable teacher model (e.g., GPT-4o) to synthesize a larger, more diverse dataset of full reasoning traces $e_t$ corresponding to various observations $o_t$.

\textbf{Distillation to Planning Thoughts:} To mitigate the context length problem, each verbose reasoning trace $e_t$ is distilled by the teacher model into a concise short planning thought $p_t$. This summary captures the essential intent and next step without extensive token overhead.

\textbf{Quaternion Assembly:} The final data structure is the planning quaternion, $Q_t = (o_t, e_t, p_t, a_t)$, which encapsulates the observation, the full reasoning trace, the distilled planning thought, and the resulting action. This structure provides a principled representation that disentangles deep, immediate deliberation from the high-level planning intent required for long-horizon consistency.

\subsubsection{Long-Short Chain-of-Thought Fusion for Efficient Deliberation}
To efficiently operationalize the planning quaternions during inference, we employ a \textit{long-short chain-of-thought fusion} strategy. At each decision step $t$, the agent first generates a full reasoning trace $e_t$ (the ``long chain'') to deliberate on the immediate problem. Subsequently, it generates a compressed planning thought $p_t$ (the ``short chain''). For all subsequent steps, only the sequence of short thoughts $\{p_1, \dots, p_{t-1}\}$ is retained in the context history. This method allows the model to benefit from detailed reasoning in the immediate step while preventing context overload and preserving the continuity of the high-level plan.

\subsubsection{Multi-Round Interactive Supervised Fine-tuning}
Our agent is trained on the generated quaternions using a step-level, multi-round interactive supervised fine-tuning objective. The introduction of the planning quaternion refines our task formulation. The agent's policy $\pi_\theta$ now generates a triplet comprising the full reasoning trace $e_t$, the concise planning thought $p_t$, and the action $a_t$. Critically, to maintain contextual efficiency, the policy is conditioned not on the full reasoning history, but on the sequence of concise planning thoughts. The generative process is thus redefined as:
\[
\begin{aligned}
[e_t, p_t, a_t] \sim \pi_\theta(&u, (o_1, p_1, a_1), \dots, \\
                                &(o_{t-1}, p_{t-1}, a_{t-1}), o_t).
\end{aligned}
\]
Consequently, the supervised fine-tuning objective is to maximize the likelihood of generating the target triplet $(e_N, p_N, a_N)$ from an expert trajectory $\mathcal{T}$, now structured as:
\begin{equation} \label{eq:trajectory_quaternion}
\begin{aligned}
\mathcal{T} = [u,\ &(o_1, p_1, a_1), \dots,\\
              &(o_{N-1}, p_{N-1}, a_{N-1}),\\
              &o_N \rightarrow (e_N, p_N, a_N)].
\end{aligned}
\end{equation}

This procedure trains the agent to perform both the reasoning and summarization required for efficient and effective long-horizon planning.

\begin{table*}[t]
    \centering
    \caption{Comparative analysis of Success Rate (\%) on the ALFWorld, ScienceWorld, and WebShop benchmarks. The proposed framework, applied to an 8B parameter model, establishes a new state-of-the-art, outperforming both fine-tuned baselines and larger proprietary reasoning models. }
    \label{tab:main_results}
    \scalebox{0.85}{
    \renewcommand{\arraystretch}{1.02}
    \begin{tabular}{l c c c c c c c}
        \toprule
        \multirow{2}{*}{\textbf{Method}} 
        & \multirow{2}{*}{\textbf{Approach}} 
        & \multicolumn{2}{c}{\textbf{ALFWorld}} 
        & \multicolumn{2}{c}{\textbf{ScienceWorld}} 
        & \multirow{2}{*}{\textbf{WebShop}}
        & \multirow{2}{*}{\textbf{Average}} \\
        \cmidrule(lr){3-4} \cmidrule(lr){5-6}
        & & Seen & Unseen & Seen & Unseen & & \\
        \midrule
        \midrule
        \multicolumn{8}{l}{\textit{Zero-shot Methods}} \\
        \midrule
        \multicolumn{8}{l}{\textit{Base Models}} \\
        Llama-3.1-8B-Instruct~\cite{dubey2024llama}   & System-1 & 32.90 & 40.30 & 42.27 & 46.92 & 60.00 & 44.88 \\
        Qwen-2.5-7B-Instruct~\cite{qwen25}    & System-1 & 72.10 & 76.10 & 28.35 & 33.18 & 91.50 & 60.65 \\
        \midrule
        \multicolumn{8}{l}{\textit{Reasoning LLMs}} \\
        o3-mini~\cite{openai2025o3mini}                 & System-2 & 58.90 & 62.70 & 56.28 & 55.45 & 76.50 & 61.57 \\
        Deepseek-R1~\cite{deepseekr1}             & System-2 & 61.40 & 53.70 & 56.70 & 59.62 & 58.50 & 58.86 \\
        Qwen-3-8B-Thinking~\cite{qwen3}      & System-2 & 49.30 & 67.20 & 57.45 & 50.27 & 71.50 & 59.14 \\
        \midrule
        \midrule
        \multicolumn{8}{l}{\textit{Fine-Tuned Methods}} \\
        \midrule
        \multicolumn{8}{l}{\textit{Llama-3.1-8B-Instruct-Based}} \\
        SFT~\cite{song2024trial}                     & System-2 & 80.00 & 71.60 & 68.04 & 72.04 & 88.00 & 75.94 \\
        ETO~\cite{song2024trial}                     & System-2 & 78.60 & 71.60 & 76.29 & 77.25 & 90.00 & 78.75 \\
        MPO~\cite{xiong2025mpo}                     & System-2 & 82.90 & 78.40 & 80.91 & 77.46 & 87.50 & 81.83 \\
        \textbf{Ours} 
                                & System-2 & \textbf{87.90} & \textbf{89.60} & \textbf{83.16} & \textbf{85.15} & \textbf{97.00} & \textbf{88.16} \\
        \midrule
        \bottomrule
    \end{tabular}
    }
\end{table*}

\subsection{Stage 2: Extrapolation via Curriculum Synthesis}
To consolidate and generalize beyond the initial reasoning patterns under the seed data distribution, the second stage enhances the agent's performance on novel tasks and scenarios. We formulate this as curriculum synthesis, generating progressively harder tasks to systematically probe and expand the agent's capabilities.

\subsubsection{Generative Task Augmentation for Out-of-Distribution Exposure}
We synthetically generate new planning challenges to expose the agent to a wider variety of problems. This is achieved by prompting the teacher model to create novel \textit{task skeletons}, defined as plausible sequences of observations, planning thoughts, and actions $(o_t, p_t, a_t)_{t=1}^N$. These skeletons are designed to differ from the seed data in structure, length, and goal complexity, constituting a targeted expansion of the training distribution. Our Planning Quaternion Data Flywheel is then used to populate these skeletons by generating corresponding full reasoning traces $e_t$, resulting in a complete synthetic out-of-distribution dataset.

\subsubsection{Complexity-Stratified Curriculum Learning}
Training directly on a heterogeneous mix of simple and complex tasks can lead to instability. We therefore employ a curriculum learning strategy, where data is organized by difficulty, primarily defined by trajectory length—a reliable proxy for planning complexity. The agent is first trained on shorter tasks to build foundational skills. As training progresses, longer and more complex trajectories from the synthetic out-of-distribution dataset are gradually introduced. This staged approach stabilizes training, mitigates catastrophic forgetting, and systematically builds the agent's capacity for complex, long-horizon planning.

\subsection{Stage 3: Self-Refinement through Reward-Gated Trajectory Refinement}
The final stage leverages the sparse reward signal as an unambiguous filter for data curation and self-improvement. This process enables a virtuous cycle of self-refinement.

\subsubsection{Success-Contingent Rejection Sampling}
The core mechanism of this stage addresses the fundamental weakness of RL in sparse-reward settings. While a binary success reward is an insufficient signal for gradient-based credit assignment across a long trajectory, it is a effective, noise-free signal for classification. It cleanly separates the space of generated trajectories into successful and unsuccessful sets. We exploit this by using the trained agent to generate numerous new trajectories. The sparse binary reward then serves as a deterministic criterion for rejection sampling: only trajectories resulting in successful task completion are retained. This creates a high-quality dataset $\mathcal{T}_{\text{success}}$ composed exclusively of effective strategies, circumventing the credit assignment problem.

\subsubsection{Iterative Policy Refinement from Successful Trajectories}
Our framework enables continuous improvement through an iterative fine-tuning loop. At each iteration $k$, the current policy $\pi_{\theta_k}$ generates a new set of trajectories. The successful trajectories are identified via rejection sampling and used to augment the training dataset, creating an enriched dataset:
\[
\mathcal{D}_{k+1} = \mathcal{D}_k \cup \mathcal{T}_{\text{success}}^{(k)}.
\]
Our model is then fine-tuned on this superior dataset to produce the next-generation policy $\pi_{\theta_{k+1}}$. This process can be interpreted as a form of policy distillation, enabling the agent to internalize successful strategies discovered through exploration. Through this iterative loop, the agent progressively masters novel and effective strategies, ultimately becoming a robust and highly capable planner.

\begin{table*}[t]
    \centering
    \caption{Ablation study illustrating the incremental contributions of each stage of our BPO framework. The results validate our three-stage design, with each component proving essential for achieving state-of-the-art performance.}
    \label{tab:ablation}
    \renewcommand{\arraystretch}{1.05}
    \begin{tabular}{l c c c c c c}
        \toprule
        \multirow{2}{*}{\textbf{Method}} 
        & \multicolumn{2}{c}{\textbf{ALFWorld}} 
        & \multicolumn{2}{c}{\textbf{ScienceWorld}} 
        & \multirow{2}{*}{\textbf{WebShop}} 
        & \multirow{2}{*}{\textbf{Average}} \\
        \cmidrule(lr){2-3} \cmidrule(lr){4-5}
        & Seen & Unseen & Seen & Unseen & & \\
        \midrule
        Llama-3.1-8B-Instruct (Base)          & 32.90 & 40.30 & 42.27 & 46.92 & 60.00 & 44.88 \\
        \midrule
        + Bootstrapped Reasoning (Stage 1)          & 77.90 & 82.10 & 76.29 & 77.73 & 87.00 & 80.60 \\
        + Curriculum Synthesis (Stage 2)        & 87.10 & 83.60 & 79.38 & 79.62 & 95.50 & 85.24 \\
        + Reward-Gated Refinement (Stage 3)       & \textbf{87.90} & \textbf{89.60} & \textbf{83.16} & \textbf{85.15} & \textbf{97.00} & \textbf{88.16} \\
        \bottomrule
    \end{tabular}
\end{table*}

\section{Experiments}
We conduct a comprehensive set of experiments to empirically validate our proposed framework. Our evaluations are designed to address four key research questions that correspond to the central claims of our work:
\begin{enumerate}
    \item \textbf{Overall Efficacy:} Does our three-stage framework demonstrate effectiveness for reasoning LLMs in long-horizon, sparse-reward agentic tasks compared to existing fine-tuning methods and powerful proprietary models?
    \item \textbf{Component Contribution:} What's the incremental value of each stage in our framework? Specifically, how do Planning Quaternion Synthesis (Stage 1), Curriculum Synthesis (Stage 2), and Reward-Gated Trajectory Refinement (Stage 3) contribute to the final performance?
    \item \textbf{Framework Robustness and Generalizability:} Is our data-centric paradigm model-agnostic, and how does its performance compare to conventional reinforcement learning approaches that directly optimize on the sparse reward signal?
    \item \textbf{Reasoning Efficiency:} Does our Long-Short Chain-of-Thought Fusion strategy effectively mitigate the computational overhead of verbose reasoning, achieving a superior trade-off between performance and token efficiency?
\end{enumerate}

\subsection{Experimental Settings}

\subsubsection{Benchmarks}
We evaluate our method on three widely used benchmarks for interactive multi-round agentic planning:

\textbf{ALFWorld}~\cite{shridhar2020alfworld}: A text-based simulator for household tasks requiring multi-step reasoning to follow high-level instructions, providing a sparse, binary success reward.

\textbf{ScienceWorld}~\cite{wang2022scienceworld}: An environment focused on complex elementary science experiments that demand long-horizon planning and provides a final score upon completion.

\textbf{WebShop}~\cite{yao2022webshop}: A simulated e-commerce environment where agents must navigate a website to purchase items based on user instructions, also yielding a final score as a reward.

Following established protocols, we use the ``seen'' splits for in-distribution evaluation and the ``unseen'' splits to assess OOD generalization for ALFWorld and ScienceWorld. For WebShop, we report results on the standard test set.

\subsubsection{Training Details}
Our three-stage training process employs AdamW optimizer with stage-specific learning rates of $2 \times 10^{-5}$, $5 \times 10^{-6}$, and $1 \times 10^{-6}$, respectively. Stage 3 runs for $K$ iterations, where $K=2$ for ScienceWorld and $K=1$ for ALFWorld and WebShop. All stages are trained for 3 epochs.

\subsubsection{Baselines}
We benchmark our framework against a comprehensive set of methods:

\textbf{Zero-Shot Models:} This includes leading open-source instruction-tuned models (Llama-3.1-8B-Instruct, Qwen-2.5-7B-Instruct) and models with specialized reasoning capabilities (o3-mini, Deepseek-R1, Qwen-3-Thinking).

\textbf{Fine-Tuned Methods:} We re-implement several state-of-the-art fine-tuning methods using Llama-3.1-8B-Instruct as the backbone. These include supervised fine-tuning SFT, ETO~\cite{song2024trial}, and methods incorporating explicit planning guidance MPO~\cite{xiong2025mpo}.

\subsubsection{Evaluation Protocol}
All agents are evaluated using a ReAct-style interaction loop. For System-2 models, the inference temperature is set to 1.0 to encourage exploration, while it is fixed at 0.0 for all other models to ensure deterministic outputs. The primary evaluation metric is Success Rate (SR), with average reward reported in the Appendix.

\subsection{Main Results}
To address our first research question, we compare the end-to-end performance of our method against all baselines. As shown in Table~\ref{tab:main_results}, our approach outperforms both fine-tuned methods and powerful proprietary models across in-distribution and OOD splits. On the ScienceWorld unseen split, our model achieves a success rate of 85.15\%, surpassing the strongest fine-tuned baseline (MPO) by 7.69 percentage points. The performance gap is even more pronounced on the challenging ALFWorld unseen split, where our method attains an 89.60\% success rate, an 11.2 percentage points improvement over MPO. These results validate our core hypothesis: reframing the problem from direct policy optimization to iterative, reward-guided data curation is an effective paradigm for training robust, generalizable reasoning LLMs for long-horizon tasks. Notably, our 8B parameter model substantially outperforms much larger, dedicated reasoning models such as DeepSeek-R1, further underscoring the efficacy of our approach.

\begin{table*}[t]
    \centering
    \caption{Performance of our framework across Llama-3.1-8B and Qwen-2.5-7B backbones. Results highlight its model-agnosticism and generalizability, with substantial gains over standard fine-tuning.}
    \label{tab:llms}
    \scalebox{0.87}{
    \renewcommand{\arraystretch}{1.05}
    \begin{tabular}{l c c c c c c c}
        \toprule
        \multirow{2}{*}{\textbf{Base Model}} 
        & \multirow{2}{*}{\textbf{Method}} 
        & \multicolumn{2}{c}{\textbf{ALFWorld}} 
        & \multicolumn{2}{c}{\textbf{ScienceWorld}} 
        & \multirow{2}{*}{\textbf{WebShop}} 
        & \multirow{2}{*}{\textbf{Average}} \\
        \cmidrule(lr){3-4} \cmidrule(lr){5-6}
        & & Seen & Unseen & Seen & Unseen & & \\
        \midrule
        \midrule
        \multirow{5}{*}{\begin{tabular}[c]{@{}l@{}}Llama-3.1-8B-Instruct\\\cite{dubey2024llama}\end{tabular}} 
        & Base      & 32.90 & 40.30 & 42.27 & 46.92 & 60.00 & 44.88 \\
        & SFT~\cite{song2024trial}       & 80.00 & 71.60 & 68.04 & 72.04 & 88.00 & 75.94 \\
        & ETO~\cite{song2024trial}       & 78.60 & 71.60 & 76.29 & 77.25 & 90.00 & 78.75 \\
        & MPO~\cite{xiong2025mpo}       & 82.90 & 78.40 & 80.91 & 77.46 & 87.50 & 81.83 \\
        &  \textbf{Ours} 
                    & \textbf{87.90} & \textbf{89.60} & \textbf{83.16} & \textbf{85.15} & \textbf{97.00} & \textbf{88.16} \\
        \midrule
        \multirow{5}{*}{\begin{tabular}[c]{@{}l@{}}Qwen2.5-7B-Instruct\\\cite{qwen25}\end{tabular}}  
        & Base      & 72.10 & 76.10 & 28.35 & 33.18 & 91.50 & 60.65 \\
        & SFT~\cite{song2024trial}       & 84.30 & 80.60 & 55.67 & 63.25 & 94.50 & 75.26 \\
        & ETO~\cite{song2024trial}       & 82.10 & 76.10 & 59.28 & 65.88 & 94.00 & 75.47 \\
        & MPO~\cite{xiong2025mpo}       & 81.40 & 88.10 & 65.98 & 60.19 & 94.50 & 78.05 \\
        & \textbf{Ours} 
                    & \textbf{90.00} & \textbf{91.00} & \textbf{78.35} & \textbf{78.20} & \textbf{97.50} & \textbf{87.81} \\
        \bottomrule
    \end{tabular}
    }
\end{table*}

\subsection{Ablation Studies}
To assess the contribution of each component (RQ2), we perform cumulative ablations by progressively adding stages of our framework. As shown in Table~\ref{tab:ablation}, each stage yields a clear performance gain. Bootstrapped Reasoning provides a strong baseline using Planning Quaternion data. Curriculum-based model extrapolation further improves generalization, notably boosting the WebShop success rate on unseen tasks from 87.00\% to 95.50\%. Finally, reward-gated trajectory refinement achieves the reported SOTA performance, confirming the necessity of all components.

\subsection{Analysis and Discussion}

\subsubsection{Generalization Across Foundational Models}
To assess the model-agnostic nature of our framework (RQ3), we apply it to a different base model, Qwen-2.5-7B-Instruct. As shown in Table~\ref{tab:llms}, our method provides a substantial performance gain regardless of the underlying LLM. For example, on the ALFWorld unseen split, it elevates the Qwen base model's success rate from 76.10\% to 91.00\%, surpassing even the state-of-the-art MPO method. These results demonstrate that our data-centric paradigm is an effective and generalizable strategy for enhancing reasoning and agentic capabilities across different LLMs.

\subsubsection{Comparison with Reinforcement Learning Methods}
A core motivation of our work is the hypothesized inadequacy of conventional RL for reasoning LLMs in this problem domain (RQ3). To evaluate this, we compare our method against SFT and popular preference optimization algorithms, DPO and GRPO. As shown in Table~\ref{tab:rl}, DPO leads to performance degradation compared to SFT, while GRPO further degrades, especially on the unseen split. These results empirically suggest that sparse reward signals fail to provide a stable learning gradient for complex reasoning tasks. In contrast, our method repurposes rewards as a data filter, validating its effectiveness in this challenging regime.

\begin{table}[t]
    \centering
    \caption{Validation on ScienceWorld showing RL methods falter under sparse rewards, while ours achieves substantial gains and resolves the credit assignment challenge.}
    \label{tab:rl}
    \scalebox{0.87}{
    \renewcommand{\arraystretch}{1.05}
    \begin{tabular}{l c c}
        \toprule
        \textbf{Approach} & \textbf{Seen} & \textbf{Unseen} \\
        \midrule
        SFT                          & 76.29 & 77.73 \\
        SFT + DPO~\cite{rafailov2023direct}                    & 73.71 & 73.46 \\
        SFT + GRPO~\cite{shao2024deepseekmath}                  & 69.93 & 64.43 \\
       \textbf{Ours} 
                                     & \textbf{83.16} & \textbf{85.15} \\
        \bottomrule
    \end{tabular}
     }
\end{table}

\subsubsection{Reasoning Efficiency}
To evaluate our Long–Short Chain-of-Thought Fusion strategy (RQ4), we compare its reasoning efficiency with strong but verbose System-2 models. As shown in Table~\ref{tab:efficiency}, our agent attains a markedly higher success rate (83.16\%) while using an order of magnitude fewer reasoning tokens (112) than DeepSeek-R1 (620) and Qwen-3-Thinking (763). These results demonstrate that planning quaternion training effectively reconciles deep reasoning with computational efficiency, alleviating context-length and inference-cost issues in long-horizon tasks.

\begin{table}[t]
    \centering
    \caption{Reasoning efficiency on ScienceWorld. BPO's Long-Short CoT Fusion outperforms larger models with significantly fewer tokens, mitigating context and computational overhead.}
    \label{tab:efficiency}
    \scalebox{0.9}{
    \renewcommand{\arraystretch}{1.05}
    \begin{tabular}{l c c c}
        \toprule
        \textbf{Models} & \textbf{Size} & \textbf{\# Tokens} & \textbf{SR (\%)} \\
        \midrule
        Deepseek-R1         & 671B & 620 & 56.70 \\
        Qwen-3-Thinking     & 8B   & 763 & 57.45 \\
        \textbf{Ours} 
                            & \textbf{8B} & \textbf{112} & \textbf{83.16} \\
        \bottomrule
    \end{tabular}
    }
\end{table}

\begin{figure}[t!]
\centering
\includegraphics[width=\columnwidth]{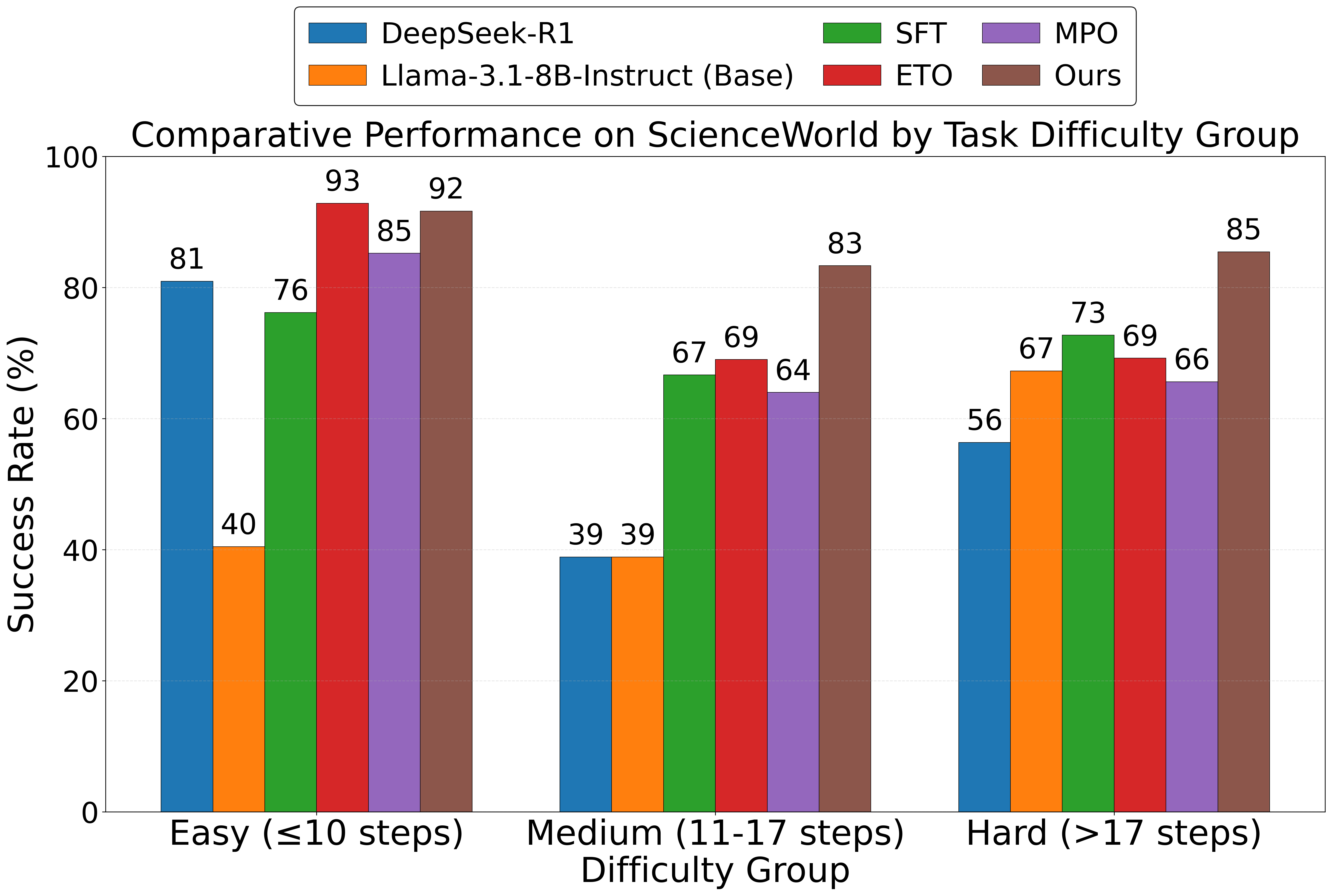}
\caption{The performance advantage of the BPO framework increases with task complexity on ScienceWorld. Tasks are grouped into Easy, Medium, and Hard based on ground-truth solution length.}
\label{fig:sciworld_difficulty_analysis}
\end{figure}

\subsubsection{Analysis of Performance by Task Complexity}
To assess robustness across task complexity, we evaluate ScienceWorld tasks grouped by solution length: Easy (\(\leq\)10 steps), Medium (11--17 steps), and Hard (\(>\)17 steps). As shown in Figure~\ref{fig:sciworld_difficulty_analysis}, our performance advantage grows with difficulty. While competitive on Easy tasks, our method achieves clear gains on Medium and Hard tasks, surpassing the next-best baselines by 14 and 12 percentage points, respectively. In contrast, most baselines degrade sharply as complexity increases, whereas our model maintains a high and stable success rate, demonstrating strong robustness in long-horizon planning.

\section{Related Work}
Our study builds upon two primary lines of research: (i) \emph{reasoning in large language models} and (ii) \emph{planning and learning for LLM-based agents in interactive environments}.
\subsection{Reasoning in Large Language Models}
Recent “System-2” models, such as OpenAI’s \textit{o1}~\cite{jaech2024openai} and DeepSeek-R1~\cite{deepseekr1}, achieve expert-level performance on complex static tasks like mathematics and programming~\cite{wei2025swe, chen2025evaluating}. Prior surveys~\cite{ke2025survey, patil2025advancing} identify two dominant paradigms: \textbf{inference-time reasoning} and \textbf{learning-to-reason}.

\paragraph{Inference-time Reasoning.}
Without updating model weights, inference-time methods allocate extra computation during inference, including multi-step prompting~\cite{sanh2021multitask, mishra2021cross}, problem decomposition~\cite{zhou2022least, khot2022decomposed}, search-based reasoning with MCTS~\cite{wang2025mcts}, and voting schemes such as self-consistency~\cite{wang2023selfconsistency}. Composite frameworks adaptively combine these techniques~\cite{liu2025bag}, while \textit{RL-of-Thoughts} trains a lightweight policy to orchestrate them~\cite{hao2025rlot}. However, these approaches incur substantial compute and context-length overhead, which our long–short chain-of-thought fusion directly alleviates.

\paragraph{Learning to Reason.}
This paradigm embeds reasoning into model parameters, typically via automatically generated supervision due to limited human-annotated chains of thought. For instance, \citet{li2025demonstrations} show that 32B models fine-tuned on 
17k distilled examples nearly match \textit{o1} on difficult math tasks. Reinforcement learning methods such as GRPO~\cite{shao2024deepseekmath} improve single-turn reasoning but struggle in multi-turn, long-horizon settings with sparse rewards~\cite{feng2025group}. Our work instead adopts a data-centric strategy, curating high-quality trajectories via rejection sampling using terminal rewards only, thereby sidestepping credit assignment issues.

\subsection{Planning and Learning in LLM-Based Agents}
Using LLMs as autonomous agents raises challenges in \emph{planning fidelity} and \emph{learning from sparse feedback}.

\paragraph{Implicit vs. Explicit Planning.} ReAct-style agents interleave reasoning and acting in a single generation pass~\cite{yao2022react}, but their plans can be myopic or prone to hallucination~\cite{zhu2024pollmgraph}. Explicit approaches therefore separate planning from execution, either by consulting external knowledge bases~\cite{guan2024} or by integrating formal planning abstractions~\cite{xiong2025mpo}. Other methods synthesize high-level outlines to guide subsequent actions~\cite{jiang2024self}; the recent Plan-and-Act framework makes this separation explicit and revisable~\cite{erdogan2025planact}.

\paragraph{Learning from Sparse Feedback.}
Environments such as ALFWorld~\cite{shridhar2020alfworld} and ScienceWorld~\cite{wang2022scienceworld} provide rewards only at task completion, limiting gradient-based RL. Prior work redistributes sparse rewards via process-level supervision~\cite{choudhury2025process, wang2025spa} or verbal feedback~\cite{shinn2024reflexion}. In contrast, we treat terminal rewards as a strict data-selection signal, iteratively training on successful trajectories and eliminating the need for dense reward shaping in long-horizon tasks.

\section{Conclusion}
In this work, we tackle two fundamental challenges in long-horizon agentic reasoning with large language models: effective credit assignment and excessive inference-time reasoning overhead. We propose a three-stage training framework that progressively bootstraps efficient reasoning behaviors, extrapolates them through synthetic curriculum generation, and further enables self-refinement via reward-gated trajectory selection. Extensive experiments demonstrate that our approach achieves state-of-the-art performance on ALFWorld, ScienceWorld, and WebShop, consistently outperforming prior methods as well as larger proprietary models, while requiring substantially fewer inference tokens. Future work will be aimed at extending the approach to multi-modal domains.  

\section{Limitations}
Our current study is restricted to text-based long-horizon benchmarks with discrete action spaces. Although the proposed framework is general in principle, extending BPO to environments with richer sensory observations, continuous control, or substantially different interaction dynamics will likely require modifications to the curriculum design and trajectory curation procedures. We leave a systematic investigation of these settings, including multimodal and continuous-control environments, as an important direction for future work.

\section*{Acknowledgments}
We sincerely appreciate our collaborators for their valuable discussions and support throughout this work. We also thank the reviewers for their insightful comments and constructive suggestions, which have improved the quality of this paper.

\bibliography{main}

\appendix

\section{Reward Comparison}
\begin{table*}[h]
    \centering
    \caption{Main results evaluated by Average Reward. These results corroborate the primary findings, indicating that our BPO framework not only achieves higher task completion but also generates more optimal action trajectories. }
    \label{tab:main_results_reward}
    \scalebox{0.85}{
    \renewcommand{\arraystretch}{1.05}
    \begin{tabular}{l c c c c c c c}
        \toprule
        \multirow{2}{*}{\textbf{Method}} 
        & \multirow{2}{*}{\textbf{Approach}} 
        & \multicolumn{2}{c}{\textbf{ALFWorld}} 
        & \multicolumn{2}{c}{\textbf{ScienceWorld}} 
        & \multirow{2}{*}{\textbf{WebShop}}
        & \multirow{2}{*}{\textbf{Average}} \\
        \cmidrule(lr){3-4} \cmidrule(lr){5-6}
        & & Seen & Unseen & Seen & Unseen & & \\
        \midrule
        \midrule
        \multicolumn{8}{l}{\textit{Zero-shot Models}} \\
        \midrule
        \multicolumn{8}{l}{\textit{Base Models}} \\
        Llama-3.1-8B-Instruct~\cite{dubey2024llama}   & System-1 & 32.90 & 40.30 & 26.64 & 28.18 & 33.93 & 32.79 \\
        Qwen-2.5-7B-Instruct~\cite{qwen25}    & System-1 & 72.10 & 76.10 & 26.68 & 27.32 & 54.28 & 51.70 \\
        \midrule
        \multicolumn{8}{l}{\textit{Reasoning LLMs}} \\
        o3-mini~\cite{openai2025o3mini}                 & System-2 & 58.90 & 62.70 & 56.95 & 54.55 & 49.02 & 56.42 \\
        Deepseek-R1~\cite{deepseekr1}             & System-2 & 61.40 & 53.70 & 63.96 & 61.13 & 40.31 & 56.50 \\
        Qwen-3-Thinking~\cite{qwen3}         & System-2 & 49.30 & 67.20 & 52.05 & 46.99 & 42.16 & 51.14 \\
        \midrule
        \midrule
        \multicolumn{8}{l}{\textit{Fine-Tuned Methods}} \\
        \midrule
        \multicolumn{8}{l}{\textit{Llama-3.1-8B-Instruct-Based}} \\
        SFT~\cite{song2024trial}                     & System-2 & 80.00 & 71.60 & 58.82 & 51.97 & 52.94 & 63.07 \\
        ETO~\cite{song2024trial}                     & System-2 & 78.60 & 71.60 & 65.69 & 58.16 & 52.08 & 65.63 \\
        MPO~\cite{xiong2025mpo}                     & System-2 & 82.90 & 78.40 & 71.61 & 53.24 & 55.20 & 68.27 \\
       \textbf{Ours} 
                                & System-2 & \textbf{87.90} & \textbf{89.60} & \textbf{77.10} & \textbf{75.67} & \textbf{67.45} & \textbf{79.14} \\
        \midrule
        \bottomrule
    \end{tabular}
    }
\end{table*}

\begin{table*}[h]
    \centering
    \caption{Ablation study results evaluated by Average Reward. These results validate the critical contribution of each stage of the BPO framework to overall agent performance and trajectory optimality. }
    \label{tab:ablation_reward}
    \renewcommand{\arraystretch}{1.05}
    \begin{tabular}{l c c c c c c}
        \toprule
        \multirow{2}{*}{\textbf{Method}} 
        & \multicolumn{2}{c}{\textbf{ALFWorld}} 
        & \multicolumn{2}{c}{\textbf{ScienceWorld}} 
        & \multirow{2}{*}{\textbf{WebShop}} 
        & \multirow{2}{*}{\textbf{Average}} \\
        \cmidrule(lr){2-3} \cmidrule(lr){4-5}
        & Seen & Unseen & Seen & Unseen & & \\
        \midrule
        Bootstrapped Reasoning (Stage 1)                 & 77.90 & 82.10 & 77.75 & 70.67 & 56.70 & 73.82 \\
       + Adversarial Curriculum (Stage 2)     & 87.10 & 83.60 & 76.01 & 74.43 & 65.75 & 77.78 \\
           + Reward-Gated Refinement (Stage 3)      & \textbf{87.90} & \textbf{89.60} & \textbf{77.10} & \textbf{75.67} & \textbf{67.45} & \textbf{79.14} \\
        \bottomrule
    \end{tabular}
\end{table*}

\begin{table*}[h]
    \centering
    \caption{Model-agnosticism analysis evaluated by Average Reward. Complementing the Success Rate results, this data confirms that the BPO framework delivers superior performance and more optimal trajectories regardless of the underlying LLM architecture.}
    \label{tab:llms_reward}
    \scalebox{0.9}{
    \renewcommand{\arraystretch}{1.05}
    \begin{tabular}{l c c c c c c c}
        \toprule
        \multirow{2}{*}{\textbf{Base Model}} 
        & \multirow{2}{*}{\textbf{Method}} 
        & \multicolumn{2}{c}{\textbf{ALFWorld}} 
        & \multicolumn{2}{c}{\textbf{ScienceWorld}} 
        & \multirow{2}{*}{\textbf{WebShop}} 
        & \multirow{2}{*}{\textbf{Average}} \\
        \cmidrule(lr){3-4} \cmidrule(lr){5-6}
        & & Seen & Unseen & Seen & Unseen & & \\
        \midrule
        \midrule
        \multirow{5}{*}{\begin{tabular}[c]{@{}l@{}}Llama-3.1-8B-Instruct\\\cite{dubey2024llama}\end{tabular}} 
        & Base      & 32.90 & 40.30 & 26.64 & 28.18 & 33.93 & 32.79 \\
        & SFT~\cite{song2024trial}       & 80.00 & 71.60 & 58.82 & 51.97 & 52.94 & 63.07 \\
        & ETO~\cite{song2024trial}       & 78.60 & 71.60 & 65.69 & 58.16 & 52.08 & 65.63 \\
        & MPO~\cite{xiong2025mpo}       & 82.90 & 78.40 & 71.61 & 53.24 & 55.20 & 68.27 \\
        & \textbf{Ours} 
                    & \textbf{87.90} & \textbf{89.60} & \textbf{77.10} & \textbf{75.67} & \textbf{67.45} & \textbf{79.14} \\
        \midrule
          \multirow{5}{*}{\begin{tabular}[c]{@{}l@{}}Qwen2.5-7B-Instruct\\\cite{qwen25}\end{tabular}}  
        & Base      & 72.10 & 76.10 & 26.68 & 27.32 & 54.28 & 51.70 \\
        & SFT~\cite{song2024trial}       & 84.30 & 80.60 & 67.99 & 63.03 & 60.67 & 71.32 \\
        & ETO~\cite{song2024trial}       & 82.10 & 76.10 & 70.29 & 61.35 & 59.62 & 69.89 \\
        & MPO~\cite{xiong2025mpo}       & 81.40 & 88.10 & 73.02 & 66.16 & 59.19 & 73.97 \\
        & \textbf{Ours} 
                    & \textbf{90.00} & \textbf{91.00} & \textbf{77.82} & \textbf{72.96} & \textbf{65.19} & \textbf{79.39} \\
        \midrule
        \bottomrule
    \end{tabular}
    }
\end{table*}

To provide a more comprehensive evaluation, we report the average reward achieved by our method and the baselines. The average reward metric complements the success rate by reflecting partial progress or task-specific scoring mechanisms present in environments such as ScienceWorld~\cite{wang2022scienceworld} and WebShop~\cite{yao2022webshop}. As demonstrated in Tables \ref{tab:main_results_reward}, \ref{tab:ablation_reward}, and \ref{tab:llms_reward}, our method consistently achieves the highest average reward across all datasets and experimental settings. These results corroborate the conclusions drawn from the success rate data in the main paper, indicating that our agent not only completes tasks more frequently but also executes more optimal trajectories.

\begin{table*}[]
\centering
\begin{tabular}{l}
\toprule
\textbf{Agent Inference Prompt for the Sciworld Benchmark}\\
\midrule
\begin{tabular}[c]{@{}l@{}}%
\textbf{\textless{}|system|\textgreater{}}: You are a helpful assistant to do some scientific experiment in an environment.\\
In the environment, there are several rooms: kitchen, foundry, workshop, bathroom, outside,\\ living room,
bedroom, greenhouse, art studio, hallway.\\
You should explore the environment and find the items you need to complete the experiment.\\
You can teleport to any room in one step.
All containers have already been opened; you can directly\\ get items from them.\\
For each of your turns, you will be given the observation of the last turn.\\ You should choose from two actions: "Thought" 
or "Action".\\
If you choose "Thought", first think about the current condition and plan for your future actions,\\ then output:\\
Thought: your thoughts.\\
Action: your next action.\\
If you choose "Action", directly output:\\
Action: your next action.\\
Remember: only one "Action:" per response is allowed.
The available actions are:\\
open OBJ: open a container\\
close OBJ: close a container\\
activate OBJ: activate a device\\
deactivate OBJ: deactivate a device\\
connect OBJ to OBJ: connect electrical components\\
disconnect OBJ: disconnect electrical components\\
use OBJ: use a device/item\\
look around: describe the current room\\
examine OBJ: describe an object in detail\\
look at OBJ: describe a container's contents\\
read OBJ: read a note or book\\
move OBJ to OBJ: move an object to a container\\
pick up OBJ: move an object to the inventory\\
pour OBJ into OBJ: pour a liquid into a container\\
mix OBJ: chemically mix a container\\
teleport to LOC: teleport to a specific room\\
focus on OBJ: signal 
intent on a task object\\
wait: take no action for 10 steps\\
wait1: take no action for 1 step:\end{tabular} \\
\bottomrule
\end{tabular}
\caption{The complete inference-time prompt used for the ScienceWorld benchmark. This prompt defines the agent's action space and the required ReAct-style output format, grounding the agent for all experimental evaluations on this environment.}
\label{tab:prompt_sciworld}
\end{table*}

\begin{table*}
\centering
\begin{tabular}{l}
\toprule
\textbf{Agent Inference Prompt for the Alfworld Benchmark}\\
\midrule
\begin{tabular}[c]{@{}l@{}}%
\textbf{\textless{}|system|\textgreater{}}: Interact with a household to solve a task.
Imagine you are an intelligent agent\\
in a household environment and your target is to perform actions to complete the task goal.\\
At the beginning of your interactions, you will be given the detailed description of the current\\ environment
and your goal to accomplish.\\
For each of your turns, you will be given the observation of the last turn.\\
You should first think about the\\
current condition and plan for your future actions, and then output your action in this turn.\\
Your output must strictly follow this format:\\
Thought: your thoughts.\\
Action: your next action.\\
The available actions are:\\
1. go to {recep}\\
2. take {obj} from {recep}\\
3. put {obj} in/on {recep}\\
4. open {recep}\\
5. close {recep}\\
6. toggle {obj} {recep}\\
7. clean {obj} with {recep}\\
8. heat {obj} with {recep}\\
9. cool {obj} with {recep}\\
where {obj} and {recep} correspond to objects and receptacles.\\
After each turn, the environment will give you immediate feedback based on which you plan your next\\ few steps.\\
If the environment outputs "Nothing happened", it means the previous\\
action was invalid; you should try\\
more options.\\
Your response should use the following format:\\
Thought: your thoughts.\\
Action: your next action.\\
\bottomrule
\end{tabular} \\
\end{tabular}
\caption{The complete inference-time prompt used for the ALFWorld benchmark. This prompt defines the agent's action space and the required ReAct-style output format, grounding the agent for all experimental evaluations on this environment.}
\label{tab:prompt_alfworld}
\end{table*}

\begin{table*}
\centering
\begin{tabular}{l}
\toprule
\textbf{Agent Inference Prompt for the WebShop Benchmark}\\
\midrule
\begin{tabular}[c]{@{}l@{}}%
\textbf{\textless{}|system|\textgreater{}}: You are web shopping.\\
I will give you instructions about what to do.\\
You have to follow the instructions.\\
Every round I will give you an observation and a list of available actions, you have to respond with an \\ action
based on the state and instruction.\\
You can use search action if search is available.\\
You can click one...[source]
\end{tabular} \\
\bottomrule
\end{tabular}
\caption{The complete inference-time prompt used for the WebShop benchmark. This prompt defines the agent's action space and the required ReAct-style output format, grounding the agent for all experimental evaluations on this environment.}
\label{tab:prompt_webshop}
\end{table*}

\begin{table*}
\centering
\begin{tabular}{l}
\toprule
\textbf{Prompt for Reasoning Distillation}\\
\midrule
\begin{tabular}[c]{@{}l@{}}%
\textless{}|\textbf{system}|\textgreater{}: You are a writing assistant specializing in enhancing reasoning passages.\\
Your task is to improve the reasoning section found between the markers \texttt{<reasoning>}\\
and \texttt{</reasoning>}.
Transform this section into a more natural, flowing chain-of-thought while:\\
- Maintaining the same logical structure and reasoning steps\\
- Adding appropriate transition words and conversational elements, like ``Okay'', ``let me see'', ``wait'',\\
``then'', ``next'', etc.\\
- Preserving the core meaning and conclusion\\
- Keeping approximately half of the original reasoning token count;
keep it concise\\
- Ensuring the content after \texttt{</reasoning>} (the final answer) remains unchanged\\
- Output in the same format as the input\\
\textless{}|\textbf{user}|\textgreater{}: {\texttt{\{reasoning\_content\}}}
\end{tabular} \\
\bottomrule
\end{tabular}
\caption{The prompt for reasoning distillation, a key component of the Planning Quaternion Data Flywheel in Stage 1. This prompt instructs the teacher model to distill a verbose reasoning trace into a concise planning thought, creating the "short chain" of thought essential for our Long-Short Chain-of-Thought Fusion strategy.}
\label{tab:prompt_reasoning_distillation}
\end{table*}

\begin{table*}
\centering
\begin{tabular}{p{0.97\textwidth}}
\toprule
\textbf{Unified Prompt for Planning Quaternion Synthesis and OOD Data Augmentation} \\
\midrule

\textless{}|\textbf{system}|\textgreater{}: You are a writing assistant specializing in enhancing reasoning passages.\\
Given an original passage and a reasoning version as examples, enhance the new passage provided with reasoning.\\

\textless{}|\textbf{user}|\textgreater{}: 
Based on the provided history of actions and observations (\verb|"Input Trajectory"|), generate the reasoning content that leads to the final action shown in the last message of the trajectory.\\

Instructions and Guidelines:
\begin{itemize}
  \item Analyze 
Context: Understand the sequence of events and the final action.
\item Generate Reasoning Only: Your output should only be the reasoning text itself.
Do not include any other text, greetings, or explanations.
  \item Natural and Logical: Make the reasoning sound natural and logical, using conversational transitions like ``Okay'', ``Let me see'', ``So'', ``Therefore'', etc., as seen in the examples.
\item Faithful: Ensure the reasoning aligns with the context and logically supports the final action.
\item Concise: Keep the reasoning brief, similar in length to the examples.
\end{itemize}

These examples below show an input trajectory (where the final message contains only the action) and the desired reasoning content that should logically precede that action.
{\{examples\_section\}}

Input Trajectory:\\
{\{target\_messages\}}\\
\bottomrule
\end{tabular}
\caption{The unified prompt for Planning Quaternion synthesis. This versatile prompt is employed in Stage 1 to generate full reasoning traces for seed trajectories and in Stage 2 to populate the synthetically generated out-of-distribution task skeletons, ensuring stylistic consistency across all training data.}
\label{tab:prompt_planning_quaternion}
\end{table*}

\begin{table*}
\centering
\begin{tabular}{l}
\toprule
\textbf{Prompt for Synthetic OOD Task Skeleton Generation}\\
\midrule
\begin{tabular}[c]{@{}l@{}}%
\textless{}|\textbf{system}|\textgreater{}: You are an expert trajectory generator for an embodied agent interacting with an
\\ environment.
Your role is to create a new interaction trajectory based on provided examples.
Each \\trajectory must follow this JSON structure: a list of dictionaries, where each dictionary has a ``role'' \\(\verb|user| or \verb|assistant|) and ``content''.\\
The structure within the list typically follows this pattern:\\
1. \verb|{"role": "user", "content": "System prompt same as examples"}|\\
2. \verb|{"role": "assistant", "content": "OK"}|\\
3. \verb|{"role": "user", "content": "Specific task description..."}|\\
4. \verb|{"role": "assistant", "content":|\\
\verb|    "Thought: [reasoning]\nAction: [chosen action]"}|\\
5. \verb|{"role": "user", "content": "Observation: [environment response]"}|\\
6. Repeat steps 4 and 5 until the task is successfully completed.\\
\\
\textless{}|\textbf{user}|\textgreater{}: 
Now, generate a new interaction trajectory following these requirements:\\
1. Task Requirements:\\
- Create a task that is novel but related to the examples.\\
- Belongs to category {\{category\}} (Topic: {\{category\_topic\}}).\\
- 
Use unseen objects distinct from those in examples.\\
- Set difficulty: {\{difficulty\}} \\(Easy: 1–2 objects; Medium: 2–3 objects;
Hard: >3 objects, complex reasoning).\\
\\
2. Trajectory Constraints:\\
- Avoid repeating the same action consecutively.\\
- Stop efficiently once the task is completed.\\
- Target length: approximately {\{target\_length\}} total messages.\\
- Begin with system prompt, assistant ``OK'', and new user task description.\\
- Maintain the style of reasoning/actions consistent with examples.\\
- Final action should be \verb|focus on OBJ|
or \verb|wait|.\\
\\
3. Output Format:\\
- Output ONLY the JSON list for the new trajectory.\\
- Enclose within triple backticks (\verb|```json... ```|).\\
- Do not add any explanatory text before or after the JSON block.\\
\\
Example trajectories are provided below.
Each follows the specified JSON structure.\\{\{trajectory\_examples\}}
\\
\bottomrule
\end{tabular}
\end{tabular}
\caption{The prompt for synthetic out-of-distribution task skeleton generation. This prompt is the core mechanism of Stage 2 (Extrapolation), used for Generative Task Augmentation to create novel and progressively more complex tasks for the curriculum, thereby enhancing the agent's robustness and generalization capabilities.}
\label{tab:prompt_ood_task_skeleton}
\end{table*}

\section{Prompts}
To ensure the reproducibility of our work, we provide the full prompts employed at key stages of our data generation and inference pipeline. Specifically, we present: (a) the environment-specific prompts for agent inference (Tables~\ref{tab:prompt_sciworld}, \ref{tab:prompt_alfworld}, and \ref{tab:prompt_webshop}); (b) the prompt for distilling verbose reasoning into concise planning thoughts (Table~\ref{tab:prompt_reasoning_distillation}); (c) the prompt for generating novel OOD task skeletons (Table~\ref{tab:prompt_ood_task_skeleton}); and (d) the unified prompt used to both synthesize initial planning quaternions and augment the OOD skeletons with full reasoning traces (Table~\ref{tab:prompt_planning_quaternion}).

\section{Case Study}
\begin{figure*}[h!]
\centering
\includegraphics[width=\textwidth]{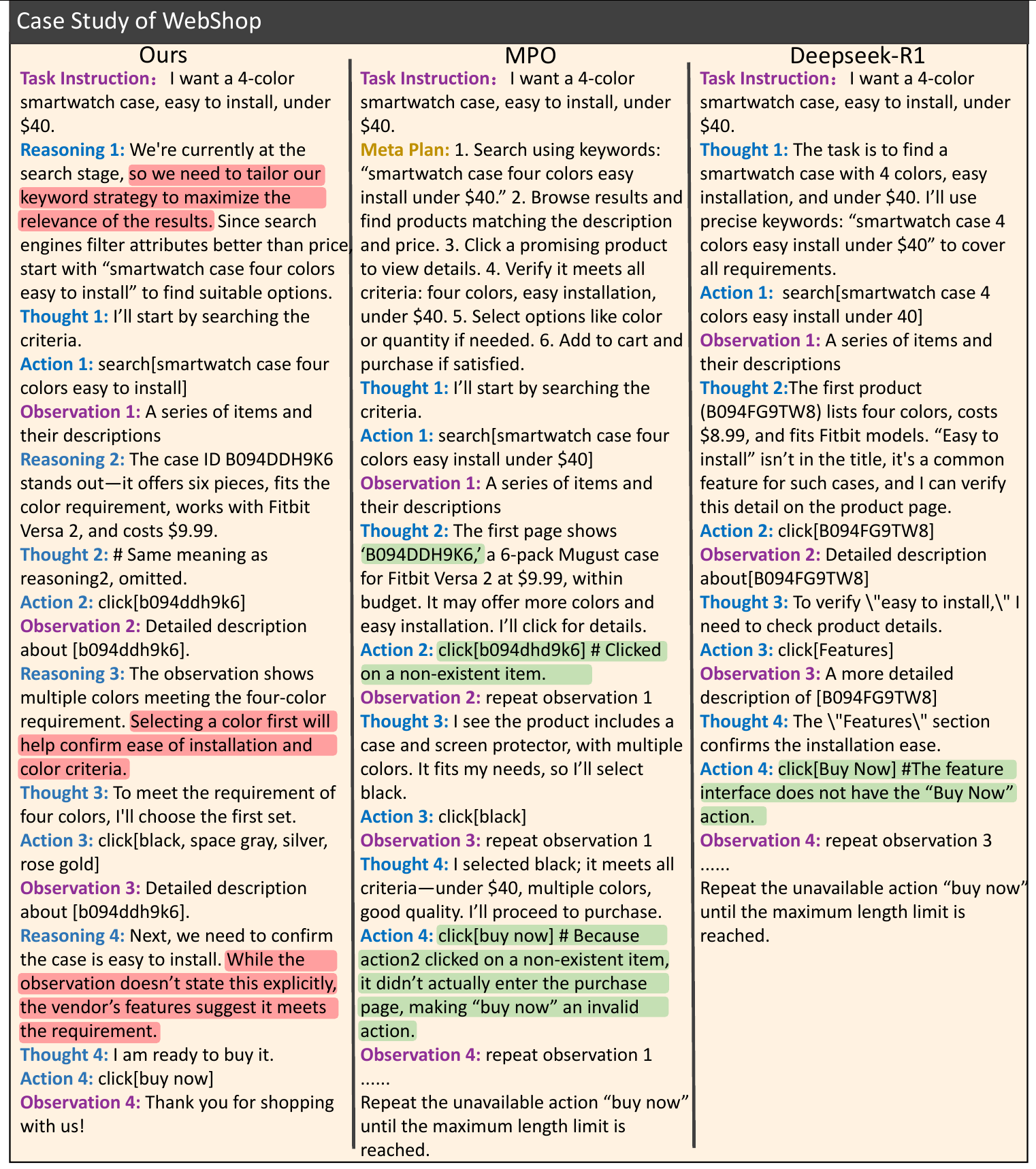}
\caption{A qualitative case study on the Webshop task “find a smartwatch case with four color options, easy installation, and under \$40.”, comparing the behavior of our agent against the MPO and Deepseek-R1 baselines. The figure contrasts our agent's successful and efficient four-step plan with the failure of the baseline methods, which ignore environmental feedback and repeatedly execute invalid actions. Red highlighting indicates our agent's effective reasoning, while green marks the incorrect decisions of the other method}
\label{fig:case_study}
\end{figure*}

We present a qualitative case study for WebShop~\cite{yao2022webshop} benchmark to highlight the differences in behavior between our agent, the MPO~\cite{xiong2025mpo} baseline, and the powerful Deepseek-R1~\cite{deepseekr1} reasoning model.
As shown in Figure~\ref{fig:case_study}, in this task, our agent correctly identified the requirements, performed deep reasoning to search for appropriate keywords, selected a suitable product, and completed the purchase—all in just four steps.
MPO, after clicking on a non-existent item, failed to correct its mistake in time. 
It ignored the observation feedback and continued executing incorrect actions as guided by its metaplan until reaching the maximum step limit, resulting in task failure.
Deepseek-R1, while demonstrating some level of deep reasoning, similarly ignored observation feedback and repeatedly performed invalid actions until reaching the maximum step limit, also leading to task failure.

\end{document}